**Deep Learning-Based Objective and Reproducible Osteosarcoma Chemotherapy Response Assessment and Outcome Prediction**

David Joon Ho[1,*], Narasimhan P. Agaram[1,*], Marc-Henri Jean[1], Stephanie D. Suser[2], Cynthia Chu[3], Chad M. Vanderbilt[1], Paul A. Meyers[2], Leonard H. Wexler[2], John H. Healey[4], Thomas J. Fuchs[5], Meera R. Hameed[1]

**Affiliations:**

[1]Department of Pathology, Memorial Sloan Kettering Cancer Center, New York, NY

[2]Department of Pediatrics, Memorial Sloan Kettering Cancer Center, New York, NY

[3]DataLine, Technology Division, Memorial Sloan Kettering Cancer Center, New York, NY

[4]Department of Surgery, Memorial Sloan Kettering Cancer Center, New York, NY

[5]Hasso Plattner Institute for Digital Health, Icahn School of Medicine at Mount Sinai, New York, NY

[*]The first two authors contributed equally

**Corresponding author:**

Meera R. Hameed, M.D.

Memorial Sloan Kettering Cancer Center

1275 York Avenue

New York, NY 10065

Tel: 212-639-5905

Email: hameedm@mskcc.org

**Conflict of Interest**

TJF is co-founder, chief scientist and equity holder of Paige.AI. CMV is a consultant (uncompensated) and equity holder in Paige.AI. DJH, NPA, CMV, TJF, and MRH have intellectual property interests related to Paige.AI, which is relevant to the work that is the subject of this paper. MSK has institutional financial interests in Paige.AI. The remaining authors declare no competing interests.

**ABSTRACT (293/300)**

Osteosarcoma is the most common primary bone cancer whose standard treatment includes pre-operative chemotherapy followed by resection. Chemotherapy response is used for predicting prognosis and further management of patients. Necrosis is routinely assessed post-chemotherapy from histology slides on resection specimens where necrosis ratio is defined as the ratio of necrotic tumor to overall tumor. Patients with necrosis ratio ≥90% are known to have better outcome. Manual microscopic review of necrosis ratio from multiple glass slides is semi-quantitative and can have intra- and inter-observer variability. In this study, we propose an objective and reproducible deep learning-based approach to estimate necrosis ratio with outcome prediction from scanned hematoxylin and eosin (H&E) whole slide images (WSIs). To conduct our study, we collected 103 osteosarcoma cases with 3134 WSIs to train our deep learning model, to validate necrosis ratio assessment, and to evaluate outcome prediction. We trained Deep Multi-Magnification Network to segment multiple tissue subtypes including viable tumor and necrotic tumor in pixel-level and to calculate case-level necrosis ratio from multiple WSIs. We showed necrosis ratio estimated by our segmentation model highly correlates with necrosis ratio from pathology reports manually assessed by experts where mean absolute differences for Grades IV (100%), III (≥90%), and II (≥50% and <90%) necrosis response are 4.4%, 4.5%, and 17.8%, respectively. Furthermore, we successfully stratified patients to predict overall survival (OS) with $p = 10^{-6}$ and progression-free survival (PFS) with $p = 0.012$. Our reproducible approach without variability enabled us to tune cutoff thresholds, specifically for our model and our data set, to 80% for OS and 60% for PFS. Our study indicates deep learning can support pathologists as an objective tool to analyze osteosarcoma from histology for assessing treatment response and predicting patient outcome. Our osteosarcoma segmentation model and code have been released at https://github.com/MSKCC-Computational-Pathology/DMMN-osteosarcoma.

# INTRODUCTION

Osteosarcoma is the most common primary bone cancer with an incidence of 4-5 cases per million worldwide in a year(1). Induction chemotherapy prior to surgery is the standard of care for osteosarcoma patients (2). Multiple studies have shown that necrosis ratio, defined as ratio of necrotic tumor to overall tumor, from histologic assessments of resected samples is one of the important prognostic factors which correlates with patient outcome (3-8). The necrosis response of tumor to chemotherapy is graded as Grade I (no or very little response), Grade II (≥50% and <90% response), Grade III (≥90% response), or Grade IV (no viable tumor) (8). The 5-year overall survival rate for patients whose necrosis ratio is greater than 90% is approximately 80%(6). However, manually assessing tumor necrosis from multiple hematoxylin and eosin (H&E)-stained slides is semi-quantitative and is prone to inter- and intra-observer variability. In one study, it was shown that necrosis ratio estimation of osteosarcoma on a H&E sectioned slide at different time points had interclass correlation coefficient of 0.652 between 6 pathologists(9).

Deep learning, a subfield of machine learning, has been widely investigated on analyzing whole slide images (WSIs) due to its nature of objectivity and reproducibility(10, 11). In osteosarcoma, multiple groups have developed deep learning models to segment viable tumor and necrotic tumor (12-16). While these models achieved acceptable performance, neither comparison with manually assessed necrosis ratio nor correlation with patient outcome data have been performed.

In this study, we propose a complete pipeline that segments multiple tissue subtypes including viable tumor and necrotic tumor in pixel-level from multiple WSIs to estimate case-level necrosis ratio in an objective and reproducible manner and to correlate the estimated necrosis ratio with overall survival and progression-free survival outcome data. Figure 1 shows the block diagram of our proposed method. For pixel-wise segmentation, we used Deep Multi-Magnification

Network(17) to accurately segment multiple tissue subtypes. From segmentation predictions of multiple WSIs, case-level necrosis ratio can be calculated by counting the number of pixels of viable tumor and necrotic tumor on WSIs. We used this data to correlate overall survival (OS) and progression-free survival (PFS). Additionally, we were able to tune the cutoff threshold to stratify patients specifically for our segmentation model and our data set. The technical details of the method and proof of concept have been previously published at a machine learning conference (18). In this work we extended our method to the largest known cohort of digital slide images from patients with osteosarcoma. The main aims of our study are: (1) to collect the largest osteosarcoma data set, (2) to develop and release a pixel-wise osteosarcoma segmentation model, (3) to estimate case-level necrosis ratio and compare with manually assessed ratio from pathologists, and (4) to correlate necrosis ratio with overall survival and progression-free survival outcome data.

**MATERIALS AND METHODS**

**Data Set**

After Institutional Review Board approval, osteosarcoma cases with resection materials available at Memorial Sloan Kettering Cancer Center (MSKCC) were selected. All cases had preoperative chemotherapy followed by resection. Detailed treatment information was available on 84 cases and the patients received combination chemotherapy including cisplatin, doxorubicin, high dose methotrexate, and/or etoposide or ifosfamide. The resected specimens are routinely sliced along the long axis and one to three representative slabs are mapped and labelled as per anatomical orientation. After routine processing, the hematoxylin and eosin-stained slides are examined microscopically for necrosis assessment (necrotic tumor divided by the overall tumor). The pathology reports were reviewed and the documented percentages of therapy-related changes were recorded. Whenever available, the follow-up data was retrieved from the clinical database. During our previous study(18), we collected 55 cases with 1578 WSIs. To increase our data set, we collected 48 additional cases with 1556 WSIs digitized in 20× magnification by Aperio AT2 scanners at MSKCC. In total, we have 103 cases with 3134 WSIs, where mean and median of the number of WSIs per case is 30.4 and 27, respectively. We used 75 WSIs from 15 training cases which were selected based on heterogeneous percentage of necrosis and the distribution of seven classes (viable tumor, necrosis with bone, necrosis without bone, normal bone, normal tissue, cartilage, and blank). We annotated a subset of whole slide images from the training cases presenting distinctive morphological patterns of the seven classes which was sufficient for the model to learn the patterns. The remaining 88 cases were used to test our segmentation model. Since pathologists microscopically review all glass slides to assess necrosis ratio, we utilized all whole slide images on testing cases to calculate necrosis ratio. First, 80 cases were used to evaluate necrosis ratio estimation after excluding 8 cases missing necrosis ratio in pathology reports. Next, we used 77 cases to predict overall survival (OS) after excluding 3 cases missing OS outcome data. Lastly, 66 cases were used to predict progression-free survival (PFS) after excluding one case missing metastasis status and 10 cases who presented with metastases at the time of diagnosis. Figure 2 shows a CONSORT flow diagram of our data set. To the best of our knowledge, this is the largest osteosarcoma data set.

**Tissue Segmentation**

Case-level necrosis ratio consists of the ratio of the area of necrotic tumor to the area of overall tumor on a set of osteosarcoma slides. Therefore, accurate pixel-wise segmentation would be necessary to count the number of pixels for viable tumor and the number of pixels for necrotic tumor on a set of osteosarcoma WSIs and to estimate the case-level necrosis ratio. WSIs are made up of giga-pixels which cannot be processed as one image due to their large size. Instead, they need to be processed in patches which are cropped square-shaped regions from the WSIs. In this study, we used Deep Multi-Magnification Network (DMMN)(17) that processes a set of patches in size of 256×256 pixels in 20×, 10×, and 5× magnifications centered at the same coordinate to accurately generate pixel-wise tissue segmentation predictions of a patch in size of 256×256 pixels in 20× magnification.

To train the segmentation model, we used Deep Interactive Learning(18) to efficiently annotate a limited set of osteosarcoma training cases. Deep Interactive Learning applies an iterative approach of correcting (or annotating) mislabeled regions from a previous model and finetuning the model with the additionally corrected patches to a training set. In this study, we finetuned the model generated in our previous work(18) segmenting seven classes including viable tumor, necrosis with bone, necrosis without bone, normal bone, normal tissue, cartilage, and blank. Specifically, we observed regions with treatment effect with an increased density of inflammatory cells, macrophages, and stromal cells were incorrectly labeled as viable tumor by the previous segmentation model. To finetune the model with these morphological patterns we included two additional cases with 26 WSI containing these patterns to the training set. Without any additional manual annotation, these mislabeled regions from the two cases were extracted in patches with the corresponding correct labels (necrosis without bone). For optimization, we used weighted cross entropy as our loss function with stochastic gradient descent with a learning rate of $5\times10^{-6}$, a momentum of 0.99, and a weight decay of $10^{-4}$ for 10 epochs. The final model was selected based on the highest mean intersection-over-union on the validation set which is a subset of the training set not used for optimization.

Since giga-pixel WSIs are too large to be segmented at once, we segmented patches starting from a window at the top, left corner of the WSIs and sliding the window to horizontal and vertical directions by 256 pixels until the entire WSIs are segmented. We did not use Otsu Algorithm(19) because we observed some necrosis regions can be excluded due to their pixel intensities. All of the implementation for training and inference was done on PyTorch(20) and all experiments were conducted on an Nvidia Tesla V100 GPU. WSIs and their segmentation predictions were visualized by our MSKCC slide viewer(21).

After all the WSIs in a case are segmented, a case-level necrosis ratio from multiple WSIs estimated by the deep learning model, $r_{DL}$, is calculated by

$$r_{DL} = \frac{p_{NT}}{p_{VT} + p_{NT}}$$

where $p_{VT}$ and $p_{NT}$ are the number of pixels for viable tumor and necrotic tumor, respectively. We compared necrosis ratio estimated by our deep learning model with necrosis ratio estimated by pathologists from pathology reports to evaluate if our segmentation model can reproduce manually assessed necrosis ratio by experts.

**Patient Stratification**

Based on necrosis ratio calculated by our segmentation model, we were able to stratify patients to predict patient outcome. We collected overall survival (OS) and progression-free survival (PFS) outcome data from patient charts and plotted Kaplan-Meier curves. Since reproducible estimation of necrosis ratio without any variability is now possible with our deep learning model, we not only tried the well-known cutoff threshold at 90%(6) but also tuned the cutoff threshold with an interval of 10% to objectively investigate various cutoff thresholds specifically for our segmentation model and our data set. The log-rank test was performed to evaluate patient stratification.

## RESULTS

**Necrosis Ratio Assessment**

Figures 3 and 4 show multi-class segmentation predictions on WSIs and zoom-in images, respectively. By overlaying the multi-class segmentation predictions on testing WSIs using our MSKCC slide viewer(21), we visually validated that our segmentation model can accurately segment seven tissue subtypes. We observed the model was not able to segment well certain morphological patterns such as isolated viable tumor cells and chondroid foci, shown in Supplementary Fig. S1.

To quantitatively evaluate our segmentation model, we compared necrosis ratio manually assessed by experts from pathology reports (denoted as $r_{PR}$) and necrosis ratio objectively assessed by our deep learning model (denoted as $r_{DL}$) using absolute difference between them. Our hypothesis is that our deep learning model would be able to calculate necrosis ratio close to the ratio manually assessed by experts. Therefore, we used absolute difference as our metric which is defined as $|r_{PR} - r_{DL}|$. Table 1 shows mean, median, and standard deviation of absolute differences in various grades where Grade IV necrosis response is defined as cases whose necrosis ratio is 100%, Grade III necrosis response as cases whose necrosis ratio is greater than or equal to 90% but less than 100%, Grade II necrosis response as cases whose necrosis ratio is greater than or equal to 50% but less than 90%, and Grade I necrosis response as cases whose necrosis ratio is less than 50% (8). Mean absolute differences for Grades IV, III, II, and I necrosis response are 4.4%, 4.5%, 17.8%, and 39.2% and median absolute differences for Grades IV, III, II, and I necrosis response are 5.9%, 2.9%, 18.4%, and 38.6%, respectively. The scatter plot of the 80 testing cases is shown in Supplementary Fig. S2. Necrosis ratio assessment shows our segmentation model can generate accurate predictions especially in Grades IV, III, and II necrosis response. We further analyzed the model using outcome data of testing cases to evaluate if our deep learning model can be clinically used, described in the next section.

**Outcome Prediction**

We plotted Kaplan-Meier curves and calculated the log-rank $p$-values to evaluate outcome predictions, shown in Fig. 5. Based on manual assessment from pathology reports at the conventional 90% cutoff threshold on 77 testing cases, we achieved $p = 0.054$ for overall survival (OS) outcome. Based on automated assessment from our deep learning model at the 90% cutoff threshold, we achieved $p = 0.0021$, showing our deep learning model can successfully stratify patients for OS outcome. Since there is no variability caused by our deep learning model, we propose an objective approach to investigate various cutoff thresholds specifically for our segmentation model and our data set. With the interval of 10%, we achieved $p = 10^{-6}$ at the 80% cutoff threshold. Furthermore, we predicted progression-free survival (PFS) outcome data on 66 testing cases using our deep learning model and achieved $p = 0.012$ at the 60% cutoff threshold. The $p$-values from various cutoff thresholds are shown in Supplementary Table S1.

**DISCUSSION**

In this study, we developed a deep learning-based approach to estimate case-level necrosis ratio from multiple hematoxylin and eosin (H&E) stained osteosarcoma whole slide images (WSIs) where necrosis ratio is known to correlate with prognosis(3-8). Specifically, we trained Deep Multi-Magnification Network(17) to objectively and reproducibly segment multiple tissue subtypes including viable tumor and necrotic tumor in pixel-level to calculate necrosis ratio. By comparing with manually assessed necrosis ratio from pathology reports, we verified that the estimation of necrosis ratio performed by our deep learning model is accurate. Furthermore, we stratified patients in overall survival (OS) and progression-free survival (PFS) based on necrosis ratio. Due to its objective manner, we were able to tune the cutoff threshold to stratify patients specifically for our trained model and our data set. In our study, the segmentation model achieved $p = 10^{-6}$ at the 80% cutoff threshold for OS and $p = 0.012$ at the 60% cutoff threshold for PFS. To our knowledge, we have conducted the first study with the largest osteosarcoma cohort to compare manually assessed necrosis ratio from pathology reports to objectively assess necrosis ratio from our deep learning model and successfully stratify patients to predict OS and PFS based on objectively assessed necrosis ratio.

High intra- and inter-observer variability of histological subtypes of in-situ and invasive cancer and necrosis percentage by manual microscopic assessment of H&E sections has been addressed in various cancer types such as lung(22, 23), breast(24), and colon(25). While necrosis ratio from histologic slides has been well-proven as a prognostic factor in osteosarcoma(3-8), this visual estimation of necrosis remains subjective(9). Even if detailed standardized diagnostic criteria are established, reducing variability would be challenging if necrosis ratio needs to be estimated from 30 or more glass slides.

Deep learning with digitized histopathology images can be used as a tool to avoid these variability(10, 11) because deep learning models can objectively and consistently generate the same output given the same input. In this study, we used Deep Multi-Magnification Network (DMMN)(17) to accurately segment viable tumor and necrotic tumor in pixel-level which is the most basic element in an image. After segmentation of osteosarcoma WSIs, we evaluated our model performance using manually assessed necrosis ratio from pathology reports and patient outcome data. To be more clinically relevant, we compared necrosis ratio from our segmentation model with necrosis ratio from pathology reports with a hypothesis that our segmentation model can reproduce necrosis ratio estimated by experts. While necrosis ratio estimated by our segmentation model highly correlates with necrosis ratio manually assessed by experts for cases with high necrosis ratio, we observed cases from Grade I necrosis response generally have higher absolute difference. Manual assessment of necrosis ratio is known to be highly subjective. For example, in their study of necrosis assessment by pathologists, Kang et al (9) showed that necrosis ratio assessed by 6 expert pathologists demonstrated interclass correlation coefficient of 0.652 for 10 cases. Additionally, high absolute differences from Grade I necrosis response may be related to imprecise subjective estimation of low percentage of necrotic tumor (<50%) which is much below the cutoff threshold (90%) used to determine good and poor prognosis (6).

Additionally, we stratified patients based on necrosis ratio estimated by our model to predict overall survival (OS) and progression-free survival (PFS) outcome data. Based on the log-rank test, we verified that our segmentation model can achieve more significant stratification than human experts. In addition, we were able to tune the cutoff threshold specifically for our segmentation model and our data set due to its objective and reproducible manner. There are some previous studies which have attempted to find the optimal cutoff threshold of necrosis ratio as a strong indicator of prognosis using manual assessment, but high intra- and inter-observer variability has precluded effective conclusions (9, 26). With a deep learning model, it would be possible to objectively and reproducibly select the optimal cutoff threshold stratifying patients with the lowest log-rank $p$-value.

There are several limitations in our study. During our qualitative evaluation of segmentation predictions, we observed our segmentation model tends to miss some viable tumor such as isolated tumor cells and chondroid foci, potentially causing overestimation of necrosis ratio. Our segmentation model was designed to segment in tissue-level, not in cell-level. Although our model can segment regions with dense areas of viable tumor cells, it misses isolated viable tumor cells because the model was not trained by cell-level annotations. Combining with a cell segmentation model(27) which can detect isolated viable tumor cells, we can further improve the estimation of necrosis ratio. Chondroid foci were underrepresented in our training set. By including more regions with rare patterns to the training set using Deep Interactive Learning(18) or generating synthetic histology images with the rare patterns using generative adversarial networks(28), we would be able to finetune the model to accurately segment them. We additionally observed artifacts caused during slide preparation (bone dust, stain precipitate) can lead to mis-segmentation which is a common challenge in all digital and computational pathology(29, 30). Training a more robust segmentation model by including artifacts in the training set would be desired. Lastly, this study was done with a data set from a single institution. For a more comprehensive study to improve segmentation and to select the optimal cutoff threshold, collecting a multi-institutional data set would be necessary.

In summary, we have conducted experiments to objectively and reproducibly estimate necrosis ratio from multiple osteosarcoma whole slide images using a deep learning-based segmentation model. Our experimental results demonstrated high correlation between manually assessed necrosis ratio by pathologists and automatically calculated necrosis ratio by our segmentation model, indicating our segmentation model can successfully estimate osteosarcoma necrosis ratio from multiple slide images. Furthermore, we were able to stratify patients to predict overall survival and progression-free survival by additionally tuning the cutoff threshold in an objective manner. As intra- and inter-observer variability is an intrinsic phenomenon in the manual and semi-quantitative estimation of necrosis ratio, adopting deep learning-based models for a more objective assessment of necrosis ratio can pave the way for more prospective studies to assess treatment response and outcome in osteosarcoma patients.

## ACKNOWLEDGEMENTS

This project was supported by the Warren Alpert Foundation Center for Digital and Computational Pathology at Memorial Sloan Kettering Cancer Center and the NIH/NCI Cancer Center Support Grant P30 CA008748. The project acknowledges support from the PRISSMM collaborative. Electronic health records were curated, and patient outcomes were defined using the PRISSMMM phenomic data system. PRISSMM is a set of phenomic data standards and tools for characterization and communication of structured information about cancer status and treatment outcomes for patients with solid tumors.

## CONFLICT OF INTEREST

TJF is co-founder, chief scientist and equity holder of Paige.AI. CMV is a consultant (uncompensated) and equity holder in Paige.AI. DJH, NPA, CMV, TJF, and MRH have intellectual property interests related to Paige.AI, which is relevant to the work that is the subject of this paper. MSK has institutional financial interests in Paige.AI. The remaining authors declare no competing interests.

## ETHICS APPROVAL AND CONSENT TO PARTICIPATE

This study was approved by the Institutional Review Board at Memorial Sloan Kettering Cancer Center (Protocol #18-013).

**AUTHOR CONTRIBUTIONS**

DJH, NPA, CMV, TJF, and MRH conceived the study. DJH developed the deep learning model. NPA and MRH reviewed and annotated whole slide images. MHJ scanned glass slides. SDS, CC, PAM, LHW, and JHH provided the clinical data set. DJH, NPA, and MRH provided statistical analysis. DJH, NPA, and MRH wrote the initial manuscript. All authors read, edited, and approved the final manuscript.

**FUNDING**

This work was supported by the Warren Alpert Foundation Center for Digital and Computational Pathology at Memorial Sloan Kettering Cancer Center and the NIH/NCI Cancer Center Support Grant P30 CA008748.

**DATA AVAILABILITY STATEMENT**

The segmentation model and code are publicly available at https://github.com/MSKCC-Computational-Pathology/DMMN-osteosarcoma. The dataset used and/or analyzed during the current study is available from the corresponding author on reasonable request.

**FIGURE LEGENDS**

**Figure 1.** Block diagram of our proposed method. Top: An osteosarcoma case with multiple slides is currently assessed via a microscope to estimate necrosis ratio and to predict outcome. Bottom: Deep learning-based segmentation by Deep Multi-Magnification Network(17) is used to segment

multiple tissue subtypes, to count the number of pixels for viable tumor (VT) and necrotic tumor (NT) to estimate necrosis ratio, and to predict outcome.

**Figure 2.** Our osteosarcoma data set containing 103 cases with 3134 whole slide images (WSIs). Fifteen cases were used to train our segmentation model and the other 88 cases were used to test the model. More specifically, 80 cases were used to evaluate necrosis ratio assessment, 77 cases were used to predict overall survival (OS), and 66 cases were used to predict progression-free survival (PFS).

**Figure 3.** Multi-class segmentation of two osteosarcoma whole slide images. Viable tumor is segmented in red, necrosis with bone in blue, necrosis without bone in yellow, normal bone in green, normal tissue in orange, cartilage in brown, and blank in gray.

**Figure 4.** Segmentation of (A,B) viable tumor, (C,D) necrosis with bone, and (E,F) necrosis without bone. Viable tumor is segmented in red, necrosis with bone in blue, necrosis without bone in yellow.

**Figure 5.** Outcome prediction. (A) Patient stratification based on overall survival (OS) outcome at the conventional 90% cutoff threshold from manually assessed pathology reports achieving p=0.054. (B) Patient stratification based on OS outcome at the same 90% cutoff threshold from our deep learning model achieving p=0.0021. The deep learning model performed a better stratification than manual assessment of glass slides. (C) Patient stratification based on OS outcome at the 80% cutoff threshold from our deep learning model achieving $p=10^{-6}$. The cutoff

threshold for our deep learning model and our data set can be tuned to have better stratification due to its objective and reproducible manner. (D) Patient stratification based on progression-free survival (PFS) outcome at the 60% cutoff threshold from our deep learning model achieving p=0.012.

**REFERENCES**

1. Ottaviani G, Jaffe N. The epidemiology of osteosarcoma. Cancer Treat Res. 2009;152:3-13.

2. Provisor AJ, Ettinger LJ, Nachman JB, Krailo MD, Makley JT, Yunis EJ, et al. Treatment of nonmetastatic osteosarcoma of the extremity with preoperative and postoperative chemotherapy: a report from the Children's Cancer Group. Journal of clinical oncology : official journal of the American Society of Clinical Oncology. 1997;15(1):76-84.

3. Davis AM, Bell RS, Goodwin PJ. Prognostic factors in osteosarcoma: a critical review. Journal of clinical oncology : official journal of the American Society of Clinical Oncology. 1994;12(2):423-31.

4. Glasser DB, Lane JM, Huvos AG, Marcove RC, Rosen G. Survival, prognosis, and therapeutic response in osteogenic sarcoma. The Memorial Hospital experience. Cancer. 1992;69(3):698-708.

5. Huvos AG, Rosen G, Marcove RC. Primary osteogenic sarcoma: pathologic aspects in 20 patients after treatment with chemotherapy en bloc resection, and prosthetic bone replacement. Archives of pathology & laboratory medicine. 1977;101(1):14-8.

6. O'Kane GM, Cadoo KA, Walsh EM, Emerson R, Dervan P, O'Keane C, et al. Perioperative chemotherapy in the treatment of osteosarcoma: a 26-year single institution review. Clinical sarcoma research. 2015;5:17.

7. Raymond AK, Chawla SP, Carrasco CH, Ayala AG, Fanning CV, Grice B, et al. Osteosarcoma chemotherapy effect: a prognostic factor. Seminars in diagnostic pathology. 1987;4(3):212-36.

8. Rosen G, Caparros B, Huvos AG, Kosloff C, Nirenberg A, Cacavio A, et al. Preoperative chemotherapy for osteogenic sarcoma: selection of postoperative adjuvant chemotherapy based on the response of the primary tumor to preoperative chemotherapy. Cancer. 1982;49(6):1221-30.

9. Kang J-W, Shin SH, Choi JH, Moon KC, Koh JS, kwon Jung C, et al. Inter-and intra-observer reliability in histologic evaluation of necrosis rate induced by neo-adjuvant chemotherapy for osteosarcoma. Int J Clin Exp Pathol. 2017;10(1):359-67.

10. Srinidhi CL, Ciga O, Martel AL. Deep neural network models for computational histopathology: A survey. Medical Image Analysis. 2021;67:101813.

11. Van der Laak J, Litjens G, Ciompi F. Deep learning in histopathology: the path to the clinic. Nature medicine. 2021;27(5):775-84.

12. Anisuzzaman D, Barzekar H, Tong L, Luo J, Yu Z. A deep learning study on osteosarcoma detection from histological images. Biomedical Signal Processing and Control. 2021;69:102931.

13. Arunachalam HB, Mishra R, Daescu O, Cederberg K, Rakheja D, Sengupta A, et al. Viable and necrotic tumor assessment from whole slide images of osteosarcoma using machine-learning and deep-learning models. PloS one. 2019;14(4):e0210706.

14. Fu Y, Xue P, Ji H, Cui W, Dong E. Deep model with Siamese network for viable and necrotic tumor regions assessment in osteosarcoma. Medical Physics. 2020;47(10):4895-905.

15. Mishra R, Daescu O, Leavey P, Rakheja D, Sengupta A, editors. Histopathological diagnosis for viable and non-viable tumor prediction for osteosarcoma using convolutional neural network. International Symposium on Bioinformatics Research and Applications; 2017: Springer.

16. Mishra R, Daescu O, Leavey P, Rakheja D, Sengupta A. Convolutional neural network for histopathological analysis of osteosarcoma. Journal of Computational Biology. 2018;25(3):313-25.

17. Ho DJ, Yarlagadda DVK, D'Alfonso TM, Hanna MG, Grabenstetter A, Ntiamoah P, et al. Deep multi-magnification networks for multi-class breast cancer image segmentation. Computerized Medical Imaging and Graphics. 2021;88:101866.

18. Ho DJ, Agaram NP, Schüffler PJ, Vanderbilt CM, Jean M-H, Hameed MR, et al., editors. Deep interactive learning: an efficient labeling approach for deep learning-based osteosarcoma treatment response assessment. International Conference on Medical Image Computing and Computer-Assisted Intervention; 2020: Springer.

19. Otsu N. A threshold selection method from gray-level histograms. IEEE transactions on systems, man, and cybernetics. 1979;9(1):62-6.

20. Paszke A, Gross S, Massa F, Lerer A, Bradbury J, Chanan G, et al. Pytorch: An imperative style, high-performance deep learning library. Advances in neural information processing systems. 2019;32.

21. Schüffler PJ, Geneslaw L, Yarlagadda DVK, Hanna MG, Samboy J, Stamelos E, et al. Integrated digital pathology at scale: A solution for clinical diagnostics and cancer research at a large academic medical center. Journal of the American Medical Informatics Association. 2021;28(9):1874-84.

22. Wang C, Durra HY, Huang Y, Manucha V. Interobserver reproducibility study of the histological patterns of primary lung adenocarcinoma with emphasis on a more complex glandular pattern distinct from the typical acinar pattern. International journal of surgical pathology. 2014;22(2):149-55.

23. Warth A, Stenzinger A, von Brünneck A-C, Goeppert B, Cortis J, Petersen I, et al. Interobserver variability in the application of the novel IASLC/ATS/ERS classification for pulmonary adenocarcinomas. European respiratory journal. 2012;40(5):1221-7.

24. Gomes DS, Porto SS, Balabram D, Gobbi H. Inter-observer variability between general pathologists and a specialist in breast pathology in the diagnosis of lobular neoplasia, columnar cell lesions, atypical ductal hyperplasia and ductal carcinoma in situ of the breast. Diagnostic pathology. 2014;9(1):1-9.

25. Viray H, Li K, Long TA, Vasalos P, Bridge JA, Jennings LJ, et al. A prospective, multi-institutional diagnostic trial to determine pathologist accuracy in estimation of percentage of malignant cells. Archives of Pathology and Laboratory Medicine. 2013;137(11):1545-9.

26. Li X, Ashana AO, Moretti VM, Lackman RD. The relation of tumour necrosis and survival in patients with osteosarcoma. International orthopaedics. 2011;35(12):1847-53.

27. Graham S, Vu QD, Raza SEA, Azam A, Tsang YW, Kwak JT, et al. Hover-net: Simultaneous segmentation and classification of nuclei in multi-tissue histology images. Medical Image Analysis. 2019;58:101563.

28. Deshpande S, Minhas F, Graham S, Rajpoot N. SAFRON: Stitching Across the Frontier Network for Generating Colorectal Cancer Histology Images. Medical Image Analysis. 2022;77:102337.

29. Janowczyk A, Zuo R, Gilmore H, Feldman M, Madabhushi A. HistoQC: an open-source quality control tool for digital pathology slides. JCO clinical cancer informatics. 2019;3:1-7.

30. Schömig-Markiefka B, Pryalukhin A, Hulla W, Bychkov A, Fukuoka J, Madabhushi A, et al. Quality control stress test for deep learning-based diagnostic model in digital pathology. Modern Pathology. 2021;34(12):2098-108.

| Grades | Range | Mean | Median | Standard Deviation | Number of cases |
| --- | --- | --- | --- | --- | --- |
| Grade IV necrosis response | $r_{PR} = 100\%$ | 4.4 | 5.9 | 2.7 | 3 |
| Grade III necrosis response | $90\% \leq r_{PR} < 100\%$ | 4.5 | 2.9 | 4.4 | 23 |
| Grade II necrosis response | $50\% \leq r_{PR} < 90\%$ | 17.8 | 18.4 | 10.0 | 19 |
| Grade I necrosis response | $0\% \leq r_{PR} < 50\%$ | 39.2 | 38.6 | 17.9 | 35 |
| All Grades | $0\% \leq r_{PR} \leq 100\%$ | 22.8 | 17.9 | 20.0 | 80 |

**Table 1**. Mean, median, and standard deviation of absolute differences on various grades based on necrosis ratio from pathology reports, denoted as $r_{PR}$.

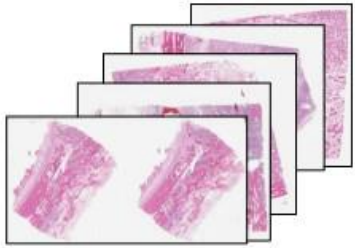
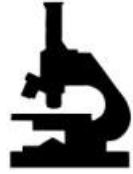
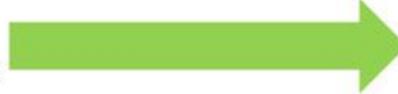
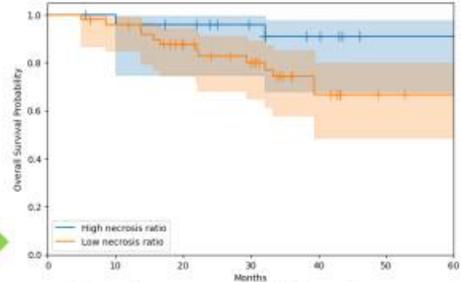

An osteosarcoma case with multiple slides

Subjective necrosis ratio assessment

Patient stratification

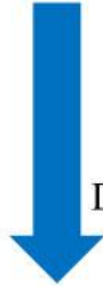
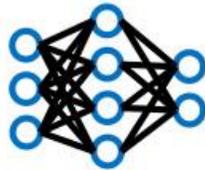
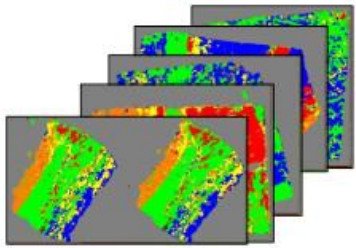
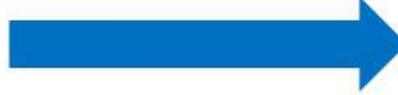
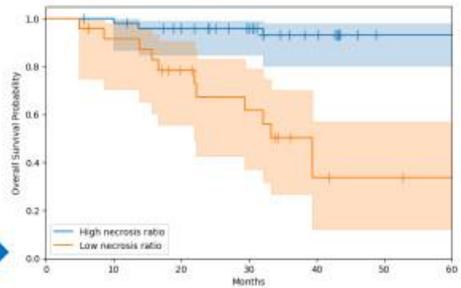

Deep Learning-Based Segmentation

Multi-class tissue segmentation

$$\frac{p_{NT}}{p_{VT} + p_{NT}}$$

Objective necrosis ratio assessment

Patient stratification

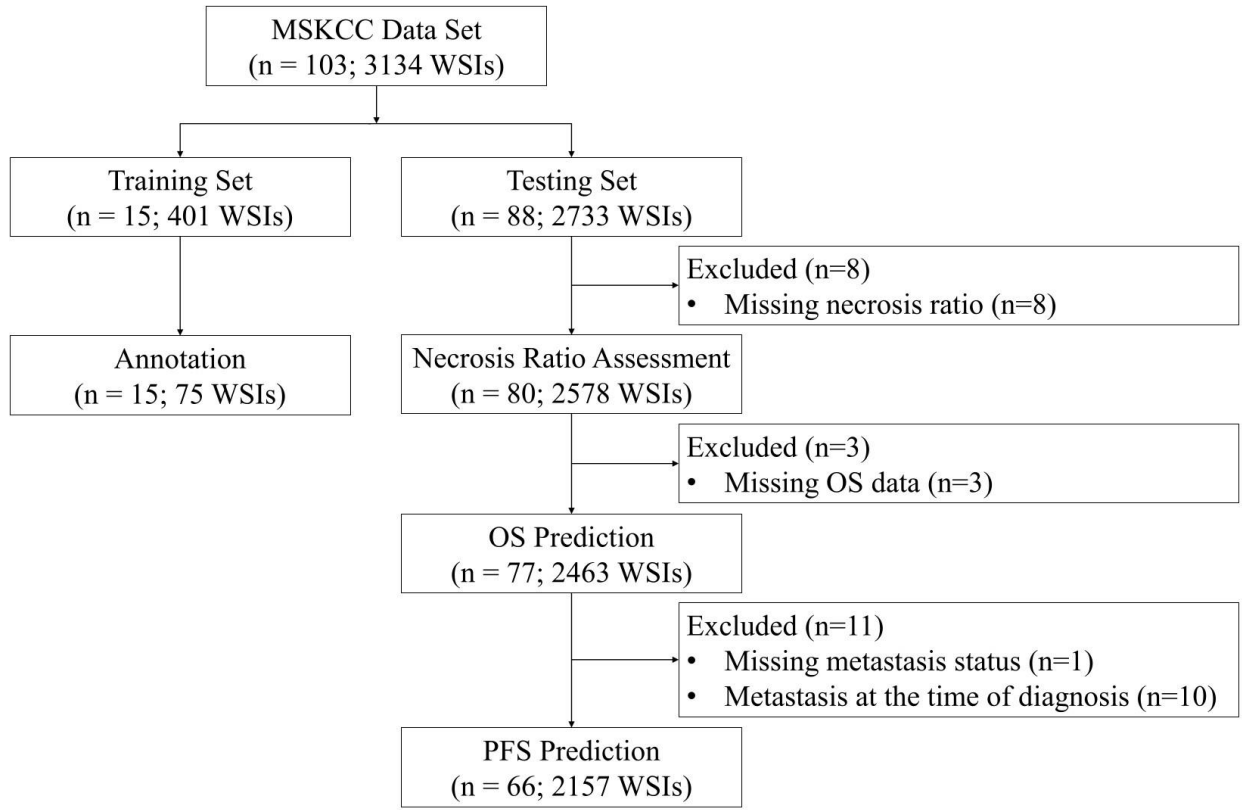

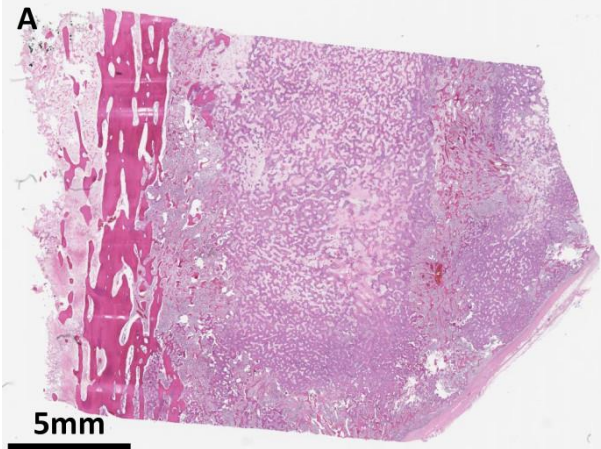
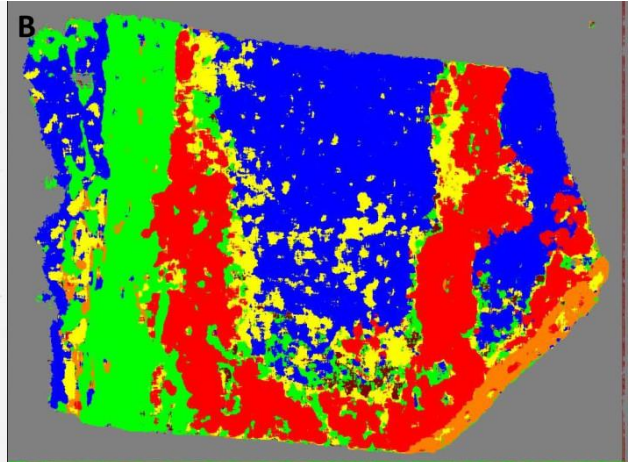
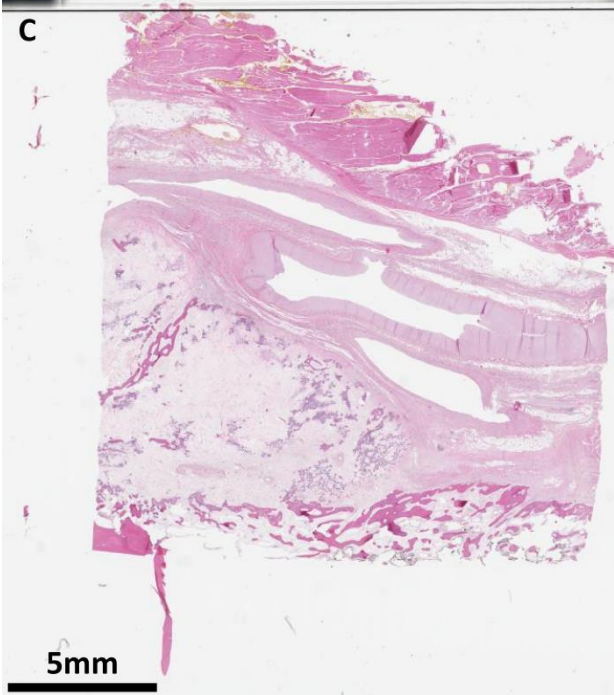
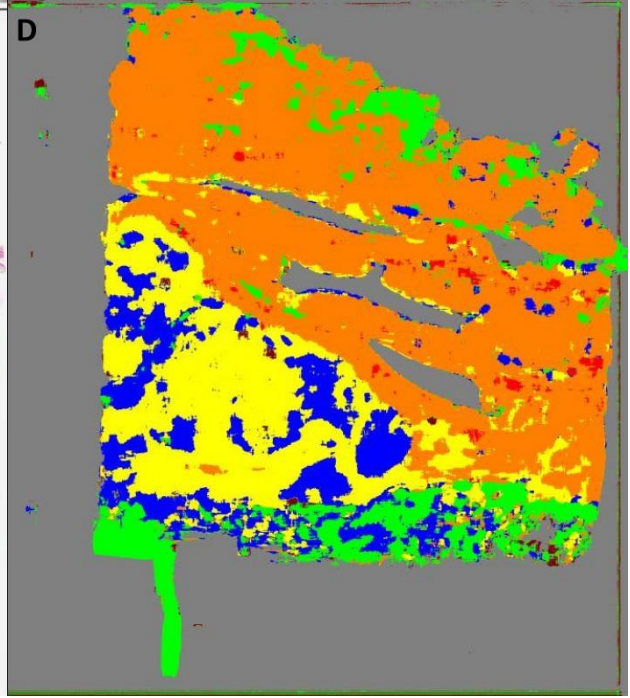

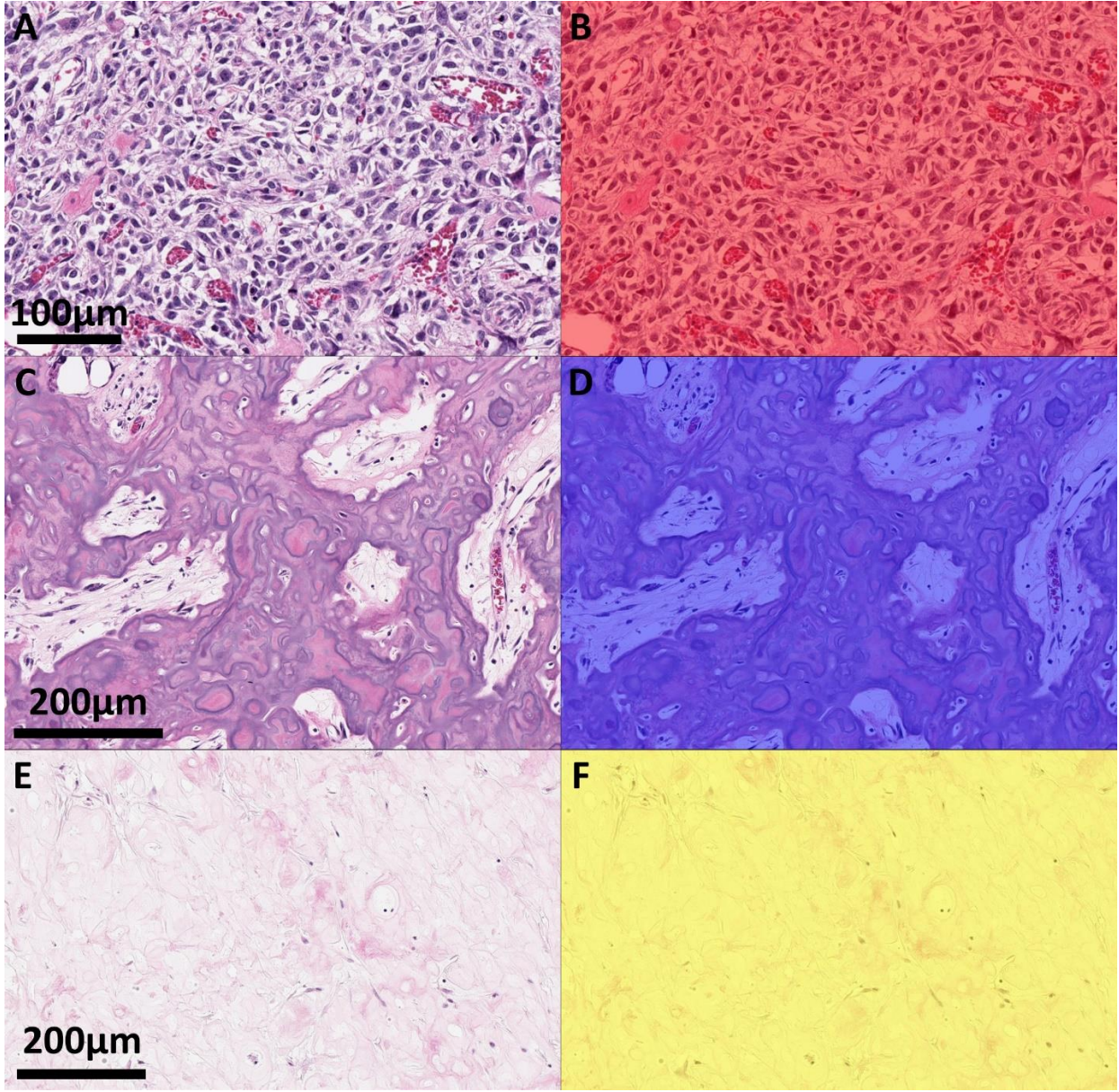

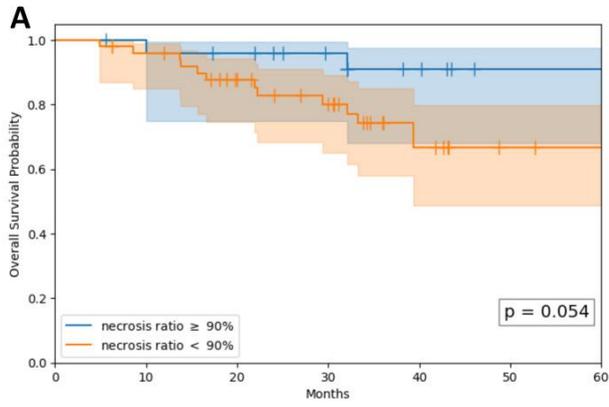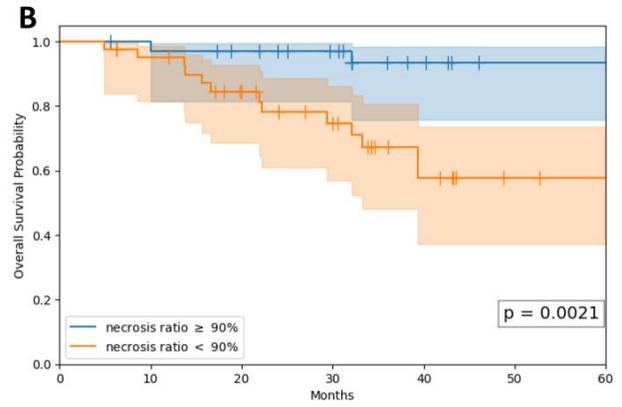
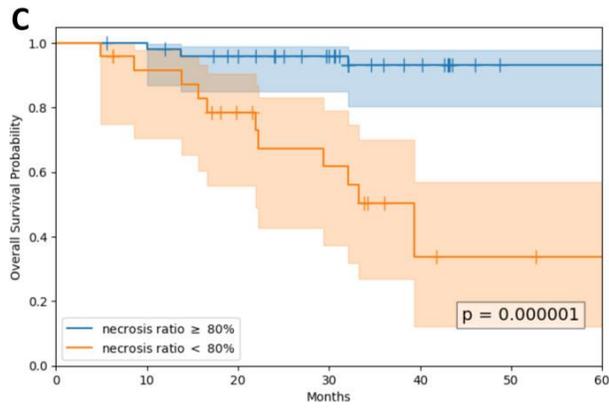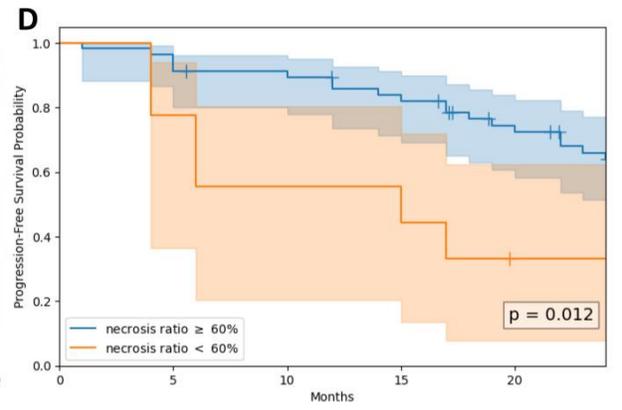

# SUPPLEMENTARY INFORMATION

**Figure S1.** Mislabeled regions from our segmentation model. (A,B) Our model was designed to segment tissue components. Although the model can segment dense viable tumor cells, it misses isolated viable tumor cells. (C,D) Our model mislabeled chondroid foci which need to be labeled as viable tumor due to the lack of sufficient examples of their morphological pattern in our training set. Regions with red, blue, yellow, green, orange, brown, gray indicates prediction of viable tumor, necrosis with bone, necrosis without bone, normal bone, normal tissue, cartilage, blank by our segmentation model, respectively.

**Figure S2.** Scatter plot between necrosis ratio from pathology reports and necrosis ratio from our deep learning model. Red, green, orange, and blue dots represent cases with Grade IV, Grade III, Grade II, and Grade I necrosis response, respectively.

| Cutoff Thresholds | Overall Survival | Progression-Free Survival |
|---|---|---|
| 90% | $2.1 \times 10^{-3}$ | 0.027 |
| 80% | **$1.4 \times 10^{-6}$** | 0.057 |
| 70% | $7.4 \times 10^{-6}$ | 0.018 |
| 60% | $3.4 \times 10^{-5}$ | **0.012** |
| 50% | $9.4 \times 10^{-3}$ | 0.072 |

**Table S1**. Log-rank p-values for overall survival (OS) and progression-free survival (PFS) outcome data with various cutoff thresholds for our segmentation model. Finding a cutoff threshold for better stratification is possible for our deep learning-based segmentation model because deep learning is objective and reproducible. The minimum p-value for OS is achieved at the 80% cutoff threshold and the minimum p-value for PFS is achieved at the 60% cutoff threshold, highlighted in bold.

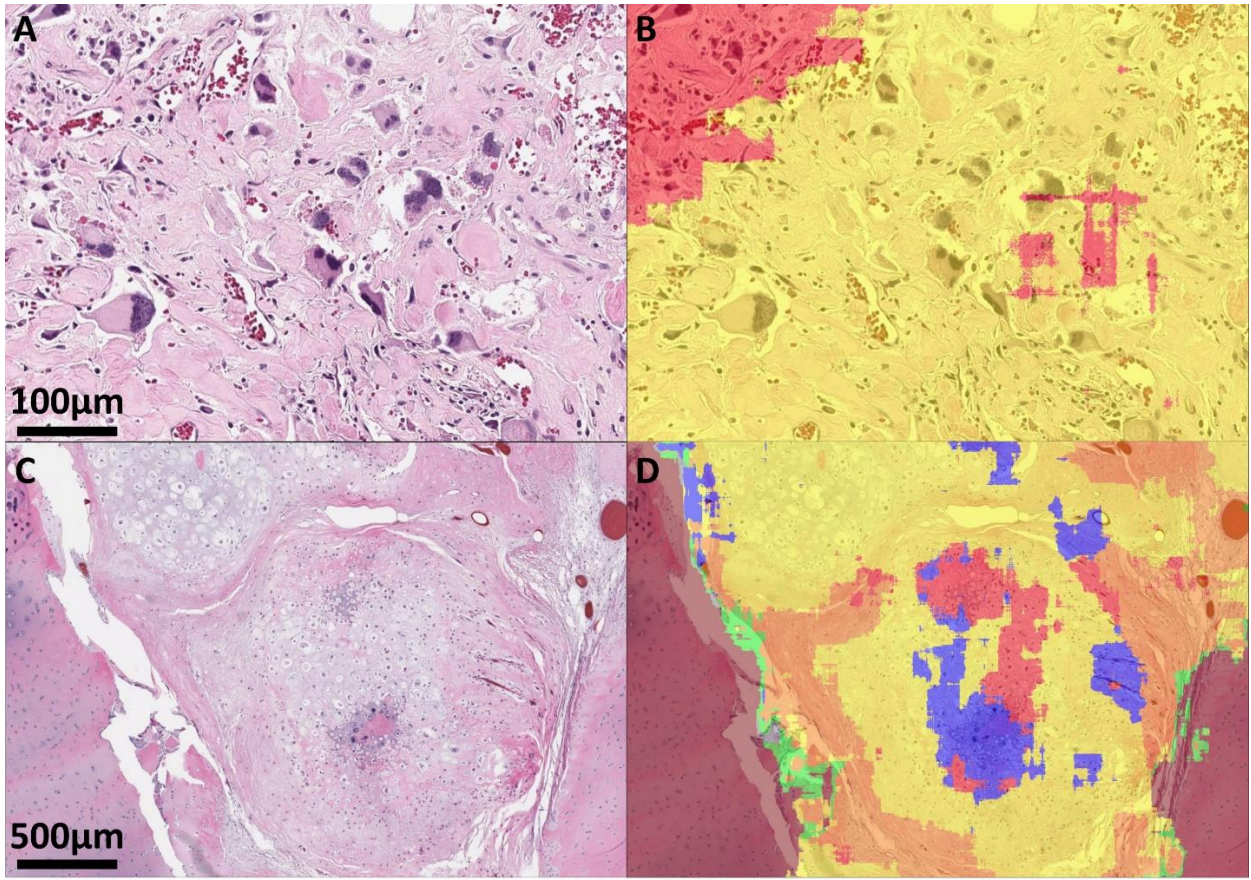

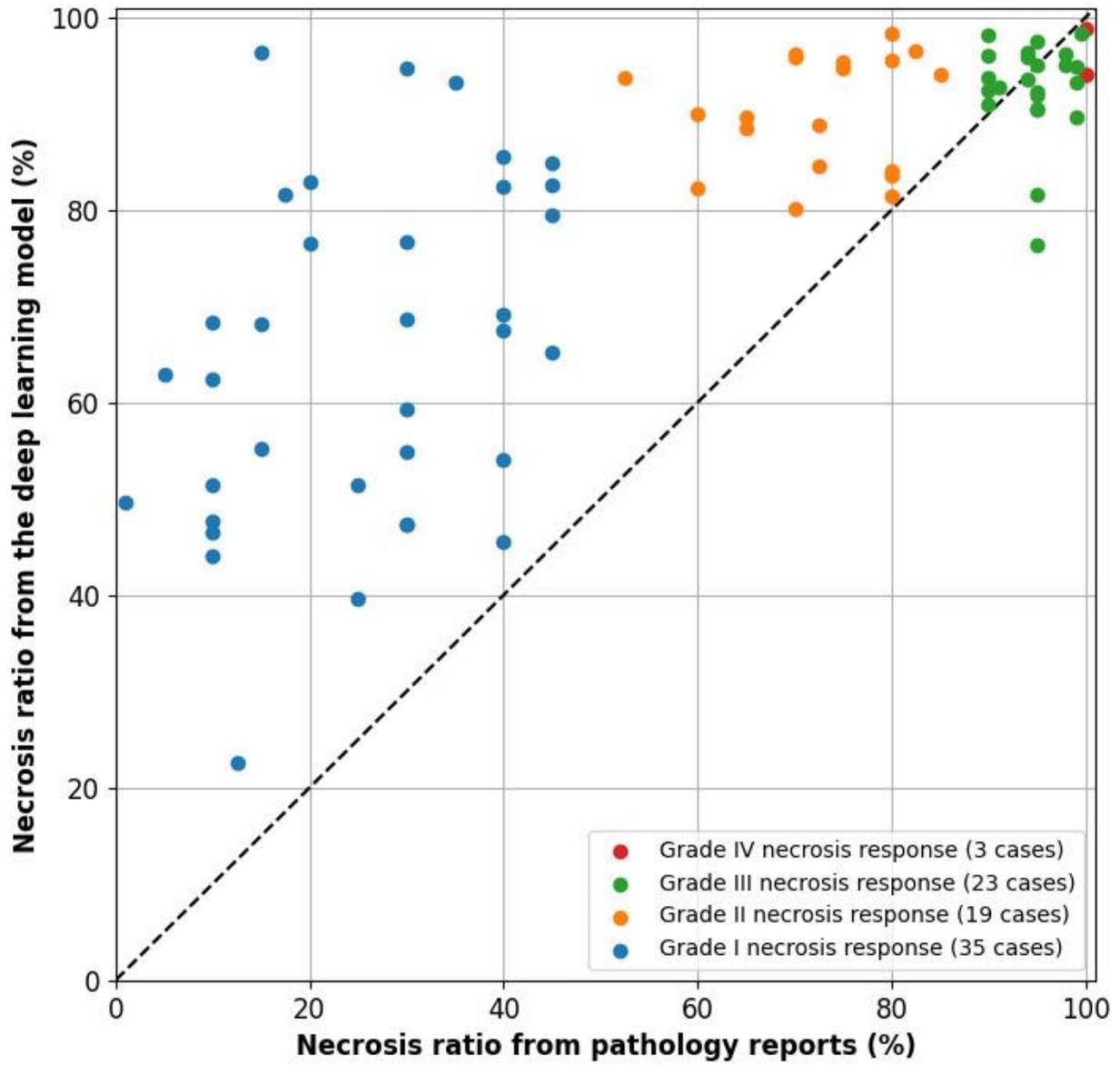